\documentclass[letterpaper]{article} % DO NOT CHANGE THIS
\usepackage{aaai2026}  % DO NOT CHANGE THIS
\usepackage{times}  % DO NOT CHANGE THIS
\usepackage{helvet}  % DO NOT CHANGE THIS
\usepackage{courier}  % DO NOT CHANGE THIS
\usepackage[hyphens]{url}  % DO NOT CHANGE THIS
\usepackage{graphicx} % DO NOT CHANGE THIS
\usepackage{natbib}  % DO NOT CHANGE THIS AND DO NOT ADD ANY OPTIONS TO IT
\usepackage{caption} % DO NOT CHANGE THIS AND DO NOT ADD ANY OPTIONS TO IT
\usepackage{algorithm}
\usepackage{algorithmic}

\usepackage{newfloat}
\usepackage{listings}
\DeclareCaptionStyle{ruled}{labelfont=normalfont,labelsep=colon,strut=off} % DO NOT CHANGE THIS
\floatstyle{ruled}
\newfloat{listing}{tb}{lst}{}
\floatname{listing}{Listing}
\usepackage{amsmath}
\usepackage{amssymb}
\usepackage{booktabs}
\usepackage{pifont}
\usepackage{subcaption}
\newcommand{\cmark}{\ding{51}}%
\newcommand{\xmark}{\ding{55}}%

\def\rvx{{\mathbf{x}}}
\def\rvX{{\mathbf{X}}}
\def\rvz{{\mathbf{z}}}
\def\rvh{{\mathbf{h}}}

\def\N{{\mathcal{N}}}
\def\E{{\mathbb{E}}}

\def\p{p_{\rm data}}
\nocopyright
\newcommand\blfootnote[1]{%
  \begingroup
  \renewcommand\thefootnote{}%
  \footnote{#1}%
  \addtocounter{footnote}{-1}%
  \endgroup
}
\title{\textbf{\textit{ShaLa}}: Multimodal \textbf{\textit{Sha}}red \textbf{\textit{La}}tent Space Modelling}
\author{
    Jiali Cui$^1$, Yan-Ying Chen$^2$, Yanxia Zhang$^2$, Matthew Klenk$^2$
}
\affiliations{
    $^1$jcui7@stevens.edu, $^2$\{yan-ying.chen, yanxia.zhang, matt.klenk\}@tri.global
}

\begin{document}

\maketitle

\begin{abstract}
\blfootnote{Under review. This work was done during Jiali Cui internship.} This paper presents a novel generative framework for learning shared latent representations across multimodal data. Many advanced multimodal methods focus on capturing all combinations of modality-specific details across inputs, which can inadvertently obscure the high-level semantic concepts that are shared across modalities. Notably, Multimodal VAEs with low-dimensional latent variables are designed to capture shared representations, enabling various tasks such as joint multimodal synthesis and cross-modal inference. However, multimodal VAEs often struggle to design expressive joint variational posteriors and suffer from low-quality synthesis. In this work, ShaLa addresses these challenges by integrating a novel architectural inference model and a second-stage expressive diffusion prior, which not only facilitates effective inference of shared latent representation but also significantly improves the quality of downstream multimodal synthesis. We validate ShaLa extensively across multiple benchmarks, demonstrating superior coherence and synthesis quality compared to state-of-the-art multimodal VAEs. Furthermore, ShaLa scales to many more modalities while prior multimodal VAEs have fallen short in capturing the increasing complexity of the shared latent space. 
\end{abstract}

% Uncomment the following to link to your code, datasets, an extended version or similar.
% You must keep this block between (not within) the abstract and the main body of the paper.
% \begin{links}
%     \link{Code}{https://aaai.org/example/code}
%     \link{Datasets}{https://aaai.org/example/datasets}
%     \link{Extended version}{https://aaai.org/example/extended-version}
% \end{links}

\section{Introduction}
\begin{figure}[!t]
    \centering
    \includegraphics[width=0.85\columnwidth]{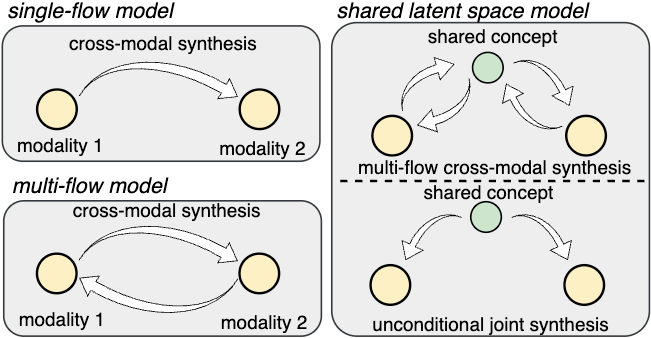}
    \caption{Comparison of multimodal modeling paradigms. \textbf{Left}: Single-flow and multi-flow models typically learn direct mappings between modalities such as images, text, or views. \textbf{Right}: Shared latent space models capture modality-invariant shared concepts, enabling both multi-flow cross-modal synthesis and unconditional joint generation. 
    %Modalities can represent general input types such as images, text, or views, \textit{etc}.
    }
\label{fig.our-illu}
\end{figure}

%Toward achieving universal AI, solving the multimodal learning problem has become a fundamental challenge and a cornerstone for building general-purpose, perceptually grounded models. In recent years, 
Deep generative models have demonstrated remarkable success in generating high-fidelity outputs across a variety of single modalities \cite{ho2020denoising,karras2020analyzing}. These advances have then rapidly extended to multimodal generation tasks, such as text-to-image, image-to-text, and other cross-modal flows \cite{ramesh2022hierarchical,alayrac2022flamingo,li2023blip}. However, many of these approaches are tailored to specific generation directions (i.e., \textit{single-flow} model) and often require dedicated models for each modality pair \cite{chen2020uniter,bao2023alluvit}. To address this limitation, recent work \cite{xu2023,le2025one} has explored unified frameworks capable of handling multiple generation flows within a single model (i.e., \textit{multi-flow} model). Yet, such models tend to emphasize capturing modality-specific combinations and fine-grained details, while often neglecting the shared semantic structure that conceptually links different modalities \cite{wu2018multimodalmvae,sutter2021} (See Fig.\ref{fig.our-illu}). 

To align with the way human cognition abstracts high-level concepts across different sensories, Multimodal Variational Autoencoders (VAEs) have emerged as a promising class of generative frameworks \cite{shi2019variationalmmvae,palumbo2023mmvae+,palumbo2024deepcmvae}. By projecting multimodal inputs into a low-dimensional shared latent space, they aim to capture modality-invariant semantic representations, forming the basis of what we refer to as shared latent variable generative modeling.

However, a major bottleneck in multimodal VAEs lies in the design of the joint inference model. Most existing approaches are built upon two dominant paradigms: Product of Experts (PoE) \cite{wu2018multimodalmvae} and Mixture of Experts (MoE) \cite{shi2019variationalmmvae}. While PoE models enforce agreement across modalities, they struggle under missing-modality scenarios. Alternatively, MoE models are more robust to missing inputs but lack the expressivity needed to accurately capture complex joint posteriors \cite{daunhawer2021limitations}. Besides this trade-off, the variational learning paradigm itself introduces an additional challenge: the prior-hole problem, a distributional mismatch between the aggregated joint posterior and the assumed prior (e.g., Gaussian or Laplacian). This gap often leads to poor sample quality and severely limits the effectiveness of multimodal VAEs in downstream generation tasks \cite{aneja2021contrastive,cui2024icml} (see more details in Sec.\ref{sec-pre-prior-hole}).

The recent success of latent diffusion models has opened new horizons for generative modeling \cite{stablediffusion,vahdat2021score}. These models are typically trained in two stages: first, a VAE is used to encode data into a compact latent space; second, a diffusion model is trained to model the aggregated posterior distribution in that space, thereby bridging the prior-posterior gap and improving synthesis quality. However, most existing works focus on single-flow mappings (e.g., text-to-image) \cite{stablediffusion} or use attention mechanisms to implicitly fuse inputs in multi-flow settings \cite{xu2023,le2025one}, without learning a shared latent variable that captures common semantic structure across modalities.

In this work, we introduce \textbf{ShaLa} — short for \textbf{Sha}red \textbf{La}tent space modeling — a generative framework for learning multimodal shared latent representation. ShaLa brings together the strengths of Multimodal VAEs and diffusion models, explicitly learning shared semantic representations through latent variables while simultaneously addressing two primary challenges: inference under missing modalities and low-quality synthesis due to the prior-hole problem.

At its core, ShaLa adopts an architectural inference model that encodes each modality into deterministic features and fuses them into a joint representation. This fused representation acts as an information bottleneck, conditioning the shared latent variable and promoting semantic alignment across modalities. On top of this, ShaLa employs a second-stage diffusion prior over the shared latent space, conditioned on these deterministic representations. The diffusion model is trained to approximate the aggregated joint posterior with modality-aware guidance, thereby overcoming the prior-hole issue and enabling robust, coherent generation. Importantly, ShaLa supports flexible inference even in the presence of missing modalities and scales to complex multi-view settings with a large number of modality instances.

In summary, our contributions include:
(i) We propose ShaLa, a novel generative framework that unifies the architectural inference and diffusion model for jointly modelling the multimodal shared latent space.
(ii) We conduct extensive experiments on standard datasets, demonstrating that ShaLa achieves state-of-the-art coherence and synthesis quality compared to other shared latent variable generative models.
(iii) We performed evaluations on the challenging multi-view dataset, which features a significantly larger number of modalities (16 views), suggesting the scalability of ShaLa for learning complex shared latent variables.

\section{Related Work}
\noindent\textbf{Multimodal Deep Generative Model.} To place multimodal VAEs in context, it is important to recognize the broader trajectory of research on multimodal generative modeling. In recent years, multimodal advances have enabled applications such as text-to-image synthesis \cite{ramesh2021zero, saharia2022photorealistic}, audio-visual generation \cite{owens2016visually, mo2023weakly}, and multi-sensory robotics \cite{lee2019making}. However, many of these models are tailored for specific cross-modal tasks (i.e., \textit{single-flow} models), typically using modality-specific architectures. To address this limitation, recent studies have proposed unified frameworks that support multiple generation flows within a single model, referred to as \textit{multi-flow} models \cite{xu2023, le2025one}. Our approach differs in that it centers on explicitly learning a shared latent space. This line of work complements existing trends in unified modeling by providing a probabilistic framework grounded in the shared representation of multimodal data. See Fig.\ref{fig.our-illu} for comparison.

\noindent\textbf{Multimodal VAE.} Multimodal VAEs have emerged as a principled and flexible class of models for learning joint representations across modalities within a shared latent space. Early efforts \cite{suzuki2016joint, hsu2018disentangling, DBLP:conf/iclr/VedantamFH018} proposed factorized latent structures and separate inference networks for each modality subset. However, such designs suffer from poor scalability, as the number of inference networks grows exponentially with the number of modalities.

To address this, MVAE \cite{wu2018multimodalmvae} introduced Product-of-Experts (PoE) formulation, which combines unimodal posteriors into a single joint distribution, enabling efficient training. This formulation was extended by MoPoE \cite{sutter2020multimodalmmJSD}, which generalized the encoder to a Mixture-of-Products, allowing richer combinations of modality subsets. Parallelly, MMVAE \cite{shi2019variationalmmvae} proposed the Mixture-of-Experts (MoE) to improve robustness for missing modalities, and follow-up work \cite{sutter2021generalizedMoPoE} proposed hybrid PoE/MoE formulations to balance expressivity and flexibility.

Other directions focused on improving the latent space itself. MVTCAE \cite{hwang2021multiMVTCAE} encouraged cross-modal consistency via total correlation regularization. MMVAE+ \cite{palumbo2023mmvae+} introduced modality-specific priors, while MVEBM \cite{yuan2024mvebm} proposed energy-based priors to capture richer structures than Gaussian prior. CMVAE (also known as D-CMVAE) \cite{palumbo2024deepcmvae} enforced clustering in the latent space for better semantic separation. Note that CMVAE integrates DiffuseVAE \cite{pandey2022} to the data generated by each modality, but the diffusion process is applied to multiple high-dimensional data outputs. Instead, ShaLa employs a diffusion process in a single low-dimensional shared latent space, allowing faster sampling and scaling to an increasing number of modalities.

Despite these advancements, a major challenge remains: most multimodal VAE variants either rely on rigid inference structures or suffer from limitations in modeling complex joint distributions, particularly in the presence of modality dropout or missing data. Our work addresses this by proposing a more expressive prior and a flexible inference structure enabled by diffusion-based modeling.

\section{Methodology}
In this section, 
%we begin by revisiting the foundations of shared latent variable generative models and Multimodal VAEs. 
we review two inference paradigms: PoE and MoE, widely adoped in Multimodal VAEs, and summarize their strengths and limitations (cf. Tab.\ref{tab.moe_vs_poe}). We then discuss the prior-hole problem, followed by our approach ShaLa.

\subsection{Preliminary: Shared Latent Variable Model} 
%We first formalize the generative modeling framework for multimodal data based on shared latent variables. 
Let $\rvX = \{\rvx_1, \rvx_2, \dots, \rvx_M\}$ represent an observation composed of $M$ modalities, where each $\rvx_i$ corresponds to one modality. The true multimodal data distribution is denoted $\p(\rvX)$, and our goal is to approximate it using a parameterized model $p_\theta(\rvX)$. Shared latent variable models introduce a global latent variable $\rvz$ to capture common semantic information across modalities, i.e., $p_\theta(\rvX)=\int p_{\theta}(\rvX, \rvz)d\rvz$.
\begin{equation}\label{joint}
\begin{aligned} 
    &p_{\theta}(\rvX, \rvz) = p_{\theta}(\rvX|\rvz)p_0(\rvz) \;\;\; \text{where}\\
    p_{\theta}(\rvX|\rvz) &= p_{\theta_{1}}(\rvx_1|\rvz)p_{\theta_2}(\rvx_2|\rvz)\cdots p_{\theta_M}(\rvx_M|\rvz)
\end{aligned}
\end{equation}
Here, $p_0(\rvz)$ is the prior distribution over the latent variable, typically assumed to be a standard Gaussian: $p_0(\rvz) = \mathcal{N}(\mathbf{0}, \mathbf{I}_d)$, where $d$ is the latent dimensionality. The conditional distribution $p_{\theta}(\rvX | \rvz)$ defines a set of modality-specific decoders parameterized by $\theta = \{\theta_1, \theta_2, \dots, \theta_M\}$, each mapping the latent code $\rvz$ to an individual modality $\rvx_M$.

To train this model, one can use maximum likelihood estimation (MLE), which maximizes the log-likelihood, i.e., $\max_\theta \mathcal{L}(\theta)=\frac{1}{n}\sum_{i=1}^N\log p_\theta(\rvX_i)$. With a sufficiently large $N$, this is equivalent to minimizing the KL divergence between the true data distribution and the model distribution, i.e., $\mathrm{KL}(\p(\rvX) || p_\theta(\rvX))$. The gradient can be written as $\frac{\partial}{\partial \theta}\mathcal{L}(\theta)=\E_{\p(\rvX)p_\theta(\rvz|\rvX)}\left[\frac{\partial}{\partial \theta}\log p_\theta(\rvX, \rvz)\right]$. However, evaluating this gradient requires access to the true joint generator posterior $p_\theta(\rvz|\rvX)$, which is typically \textit{intractable}.

\subsection{Preliminary: Multimodal VAE}
To address the intractability of the true posterior, Multimodal VAEs leverage Variational Autoencoders (VAEs) \cite{kingma2013auto} as an inference network (or encoder) and optimize a tractable surrogate objective known as the Evidence Lower Bound (ELBO). Hence, the central question becomes how to design an effective joint inference distribution $q_\phi(\rvz|\rvX)$.  Two main paradigms have emerged:

\noindent\textbf{Product-of-Experts (PoE).}\label{sec-pre-poe} The PoE  introduced by MVAE \cite{wu2018multimodalmvae}, defines the joint posterior approximation as the product of modality-specific encoders:
\begin{equation}\label{poe}
    q_\phi(\rvz|\rvX) = \prod_{i=1}^Mq_{\phi_i}(\rvz_i|\rvx_i)\nonumber
\end{equation}
Each encoder provides complementary evidence about $\rvz$, and the product enforces consistency across modalities, leading to sharp and coherent posteriors when all modalities are present. However, 
%PoE is sensitive to missing data: 
the product becomes ill-defined when one or more modalities are missing. 
%To address this, MVAE introduces a sub-sampled ELBO objective that trains the model using randomly selected modality subsets. While this improves robustness in practice, 
Therefore, PoE alone does not naturally support flexible inference under partial observations, e.g, cross-modal inference, and the training procedure remains heuristic.

\noindent\textbf{Mixture-of-Experts (MoE).}\label{sec-pre-moe} To better facilitate missing modalities, MMVAE \cite{shi2019variationalmmvae} proposes the MoE.
\begin{equation}\label{moe}
    q_\phi(\rvz|\rvX) = \frac{1}{M}\sum_{i=1}^Mq_{\phi_i}(\rvz_i|\rvx_i)\nonumber
\end{equation}
This formulation inherently supports partial input by excluding missing modalities from the mixture. However, due to its averaging nature, MoE often produces over-smoothed posteriors and weak modality alignment. As such, it may struggle to model fine-grained semantic interactions between modalities, limiting its expressiveness \cite{daunhawer2021limitations}.
\begin{table}[!t]
    \centering
    \resizebox{0.9\columnwidth}{!}{
    \begin{tabular}{c c c}
    \hline
    Method & Modelling Expressivity & Cross-modal Inference\\
    \hline
    Product of Expert  & \cmark & \xmark \\
    \hline
    Mixture of Expert  & \xmark & \cmark \\
    \hline
    \textbf{ShaLa }    & \cmark & \cmark \\
    \hline
    \end{tabular}
    }
    \caption{Comparison with PoE and MoE in terms of expressivity and support for cross-modal inference. In contrast, ShaLa facilitates both capabilities. See Sec.\ref{sec-shala-inf} and Sec.\ref{sec-shala-diff}.}
    \label{tab.moe_vs_poe}
\end{table}

\noindent\textbf{Prior-hole Problem.}\label{sec-pre-prior-hole} Another major challenge in multimodal VAEs arises from the mismatch between the aggregated posterior and the fixed prior, known as the prior-hole problem \cite{aneja2021contrastive, cui2024icml}. The aggregated posterior is defined as $q_\phi(\rvz)=\int q_\phi(\rvz|\rvX)\p(\rvX)d\rvX$. In practice, this distribution often occupies only a narrow subset of the latent space, leaving large regions of the prior unaligned with any training data. As a result, prior samples may fall into these low-density “holes” (see Fig.\ref{fig.prior-hole}), leading to low-quality generations. 
\begin{center}
    \begin{minipage}{0.99\columnwidth}
        \centering
        \includegraphics[width=0.9\columnwidth]{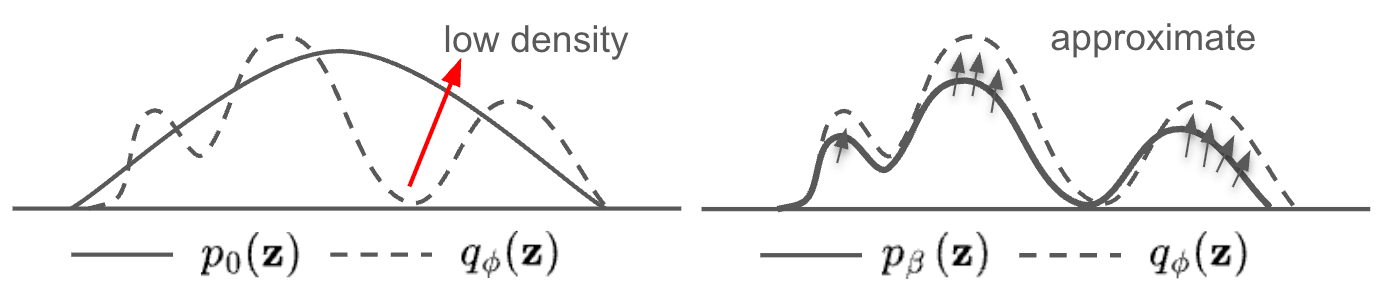}
        \captionof{figure}{Prior-hole problem: the aggregated posterior (dashed line) occupies a narrow region of the latent space, while the prior (solid line in the left figure) covers a broader area, leading to mismatched samples. Our diffusion prior (solid line in the right figure) is learned to approximate this aggregated posterior, bridging the distribution gap.}
        \label{fig.prior-hole}
    \end{minipage}
\end{center}

\begin{figure*}[!t]
    \centering
    \includegraphics[width=0.8\linewidth]{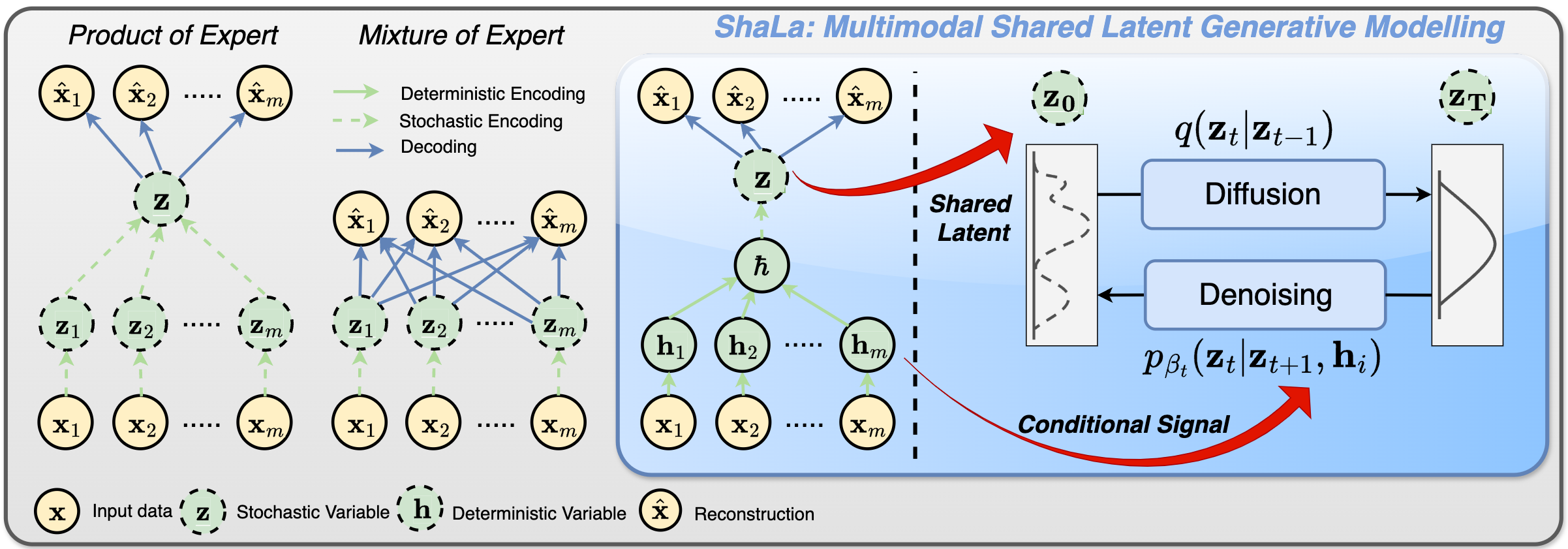}
    \caption{\textbf{Left}: PoE and MoE models construct the joint posterior by combining modality-specific \textit{stochastic} encoders, each introducing limitations under missing modalities or expressivity constraints. \textbf{Right}: ShaLa replaces stochastic encoders with \textit{deterministic} modality-specific embeddings fused into a shared representation, which then conditions a latent diffusion prior, enabling robust and flexible inference while bridging the prior-posterior gap for high-quality multimodal generation.}
    \label{fig:enter-label}
\end{figure*}
\subsection{ShaLa: Architectural Inference Model}\label{sec-shala-inf}
We now introduce ShaLa to overcome two major challenges in multimodal VAEs: (1) the design of a flexible and expressive joint inference model, and (2) the prior–posterior mismatch known as the prior-hole problem. ShaLa addresses these issues by combining a novel architectural inference model with a second-stage diffusion-based prior. 

Unlike PoE and MoE, ShaLa adopts a more direct approach by parameterizing the joint posterior $q_{\phi_i}(\rvz|\rvX)$ as a single conditional Gaussian distribution, whose parameters are inferred from the entire set of input modalities. Rather than relying on explicit unimodal posteriors, we encode each modality $\rvx_i$ into a deterministic representation $\rvh_i$ using a modality-specific encoder composed of multiple convolutional and down-sampling layers. These deterministic embeddings are then fused via a shared function $\odot(\cdot)$, implemented in practice as concatenation followed by several linear layers, into a global summary $\hbar$, which conditions the final posterior:
\begin{align}\label{our-inf}
&\rvh_1, \rvh_2, \dots, \rvh_M = I_{\phi}(\rvx_1, \rvx_2, \dots, \rvx_M)\nonumber\\
&\mathbf{\hbar} = \odot(\rvh_1, \rvh_2, \dots, \rvh_M)\\
&\rvz \sim \N(\mu_\phi(\hbar), V_\phi(\hbar))\nonumber
\end{align} 
Here, $\hbar$ serves as an information bottleneck, summarizing modality-specific semantics into a compact, high-level abstraction. By conditioning solely on this fused representation $\hbar$, our approach avoids the complexity of combining individual posteriors while retaining the ability to capture rich cross-modal interactions. Importantly, this formulation naturally supports guiding the generative process from the fused features and also forms the foundation for our diffusion-based prior in the next stage.

\subsection{ShaLa: Expressive Prior via Latent Diffusion.}\label{sec-shala-diff} 
While our architectural inference model offers a compact parameterization of the joint posterior $q_\phi(\rvz|\rvX)$, it inherently assumes access to all modalities at inference time. 
%This design lacks native support for cross-modal inference, a crucial requirement in multimodal learning where partial modality observations are common. 
In particular, since we do not model per-modality posteriors (as in PoE or MoE), our inference pathway cannot readily accommodate missing modalities at test time. 

To overcome this limitation and simultaneously address the prior-hole problem, we introduce a second-stage latent diffusion model that serves as a flexible prior over the shared latent space. This diffusion-based prior enables cross-modal inference by conditioning its generative process on deterministic modality-specific features $\rvh_i$, allowing generation from any subset of modalities without retraining the encoder.

\noindent\textbf{Forward and Reverse Process.} 
We adopt the denoising diffusion probabilistic model (DDPM) \cite{ho2020}, applied in shared latent space. Given a latent sample $\rvz_0 \sim q_\phi(\rvz)$ from the aggregated posterior, we define a forward diffusion trajectory $\rvz_{0:T} = \{\rvz_0, \rvz_1, \dots, \rvz_T\}$ using the Gaussian transition:
\begin{equation} \label{diffusion-forward}
q(\rvz_t|\rvz_{t-1}) = \mathcal{N}(\rvz_t; \alpha_t \rvz_{t-1}, \sigma_t^2 \mathbf{I})
\end{equation}
Here, $\alpha_t := \sqrt{1-\sigma^2_t}$ controls signal preservation, and $\sigma_t$ is defined by a pre-specified schedule. As $t \rightarrow T$, $\rvz_T$ approaches an isotropic Gaussian $q(\rvz_T) \sim \mathcal{N}(\mathbf{0}, \mathbf{I}_d)$. The full forward trajectory is:
\begin{equation} \label{diffusion-forward-trajectory}
q(\rvz_{0:T}) = q(\rvz_0)\prod_{t=1}^{T}q(\rvz_t|\rvz_{t-1})
\end{equation}
where $q(\rvz_{0})=q_{\phi}(\rvz)=\int q_{\phi}(\rvz|\rvX)\p(\rvX)d\rvX$. This process gradually transforms latent representations into noise, which is our prior distribution defined in Eqn.\ref{joint}.

To model the generative prior, we learn a reverse process that bridges from our prior distribution back to the aggregated posterior distribution of interest as
\begin{equation} \label{diffusion-backward}
p_{\beta_t}(\rvz_t|\rvz_{t+1}) = \mathcal{N}(\rvz_t; \mu_{\beta_t}(\rvz_{t+1}), V_{\beta_t}(\rvz_{t+1}))
\end{equation}
The full generative trajectory is
\begin{equation} \label{diffusion-backward-trajectory}
p_{\beta}(\rvz_{0:T}) = p(\rvz_T)\prod_{t=0}^{T-1}p_{\beta_t}(\rvz_t|\rvz_{t+1})
\end{equation}
This learned prior thus bridges the mismatch between the aggregated posterior and the assumed Gaussian prior, resolving the prior-hole problem and improving generation quality.

\noindent\textbf{Cross-Modal Conditioning.} 
To support cross-modal inference, i.e., sampling from $p_\beta(\rvz|\text{subset of modalities})$, we condition the reverse process on deterministic modality-specific embeddings, $\rvh_1, \dots, \rvh_M$ from Eqn.\ref{our-inf}.
%that the inference network produces representations $\rvh_1, \dots, \rvh_M$ for each modality, fused into a latent posterior via $\hbar$ ($= \odot(\rvh_1, \dots, \rvh_M)$). 
During training, we adopt a random conditioning strategy inspired by prior work \cite{zhang2023text, bie2024renaissance}, which improves flexibility in multimodal generation. Concretely, we randomly sample one available modality-specific embedding $\rvh_j \in \{\rvh_1, \dots, \rvh_M\}$ and condition the reverse step as
\begin{equation} \label{our-diffusion-backward}
\begin{aligned}
    p_{\beta_t}(\rvz_t|\rvz_{t+1}, \rvh_j) &= \mathcal{N}(\mu_{\beta_t}(\rvz_{t+1}, \rvh_j), V_{\beta_t}(\rvz_{t+1}, \rvh_j))\\
    \text{where} \quad \rvh_j &\sim \mathrm{Uniform}(\rvh_{1:M})
\end{aligned}
\end{equation}
This approach ensures that any single modality can serve as conditioning input, enabling robust inference even when other modalities are missing.

We train the diffusion prior via MLE, optimizing the marginal likelihood of latent trajectories under the learned generative model. Following common DDPM practices \cite{ho2020}, we optimize a time-averaged surrogate by randomly sampling diffusion step $t$ and computing $\frac{\partial}{\partial \beta}\E_{\text{Uni}(t),\text{Uni}(\rvh_{1:m}),q(\rvz_{0:T})}\left[\log p_{\beta_t}(\rvz_t|\rvz_{t+1}, \rvh_j)\right]$. 
%This formulation supports scalable training, reduced variance, and robust generalization, while yielding a powerful latent prior that supports high-quality multimodal generation even under partial observations.

\noindent\textbf{Joint and Cross-Modal Sampling.} Our framework supports both unconditional generation and conditional cross-modal synthesis.
During training, we randomly drop conditioning signals to enable the model to learn both conditional and unconditional pathways within a unified diffusion framework. 
This allows the model to flexibly control the strength of conditioning using a guidance scale, enabling high-quality generation under both complete and incomplete observations. We provide algorithms, derivations and implementation details in the Appendix.

\section{Experiment}
Our experimental objective is to evaluate the learned latent representation of ShaLa in terms of multimodal coherence and scalability compared to 
%We conduct extensive empirical evaluations of ShaLa across 
several established baselines of shared latent variable generative modeling. 
%Our experimental objectives include assessing ShaLa to: 
Toward that goal, we assess ShaLa in the following tasks: (1) generate semantically coherent outputs across modalities, (2) perform conditional cross-modal inference, (3) scale to many more modalities in multi-view generation tasks, and (4) maintain the quality and structure of the shared latent space. 

Additional experiments, ablation studies, and qualitative comparisons are presented in the Appendix.

\noindent\textbf{Baselines.} To ensure fair and consistent comparisons, we benchmark ShaLa against a diverse set of representative shared latent variable generative models, including MVAE \cite{wu2018multimodalmvae}, MVTCAE \cite{hwang2021multiMVTCAE}, mmJSD \cite{sutter2020multimodalmmJSD}, MoPoE \cite{sutter2021generalizedMoPoE}, MMVAE \cite{shi2019variationalmmvae}, MMVAE+ \cite{palumbo2023mmvae+}, CMVAE \cite{palumbo2024deepcmvae}, and MVEBM \cite{yuan2024mvebm}. 

\noindent\textbf{Datasets.} We utilize standard multimodal datasets used in our baselines, such as PolyMNIST \cite{sutter2021generalizedMoPoE}, MNIST-SVHN-Text (MST) \cite{sutter2020multimodalmmJSD}, and Caltech UCSD Birds (CUB) \cite{wah2011caltech}, which provide clear and interpretable benchmarks for assessing generation quality and semantic coherence across modalities. 

\begin{center}
    \begin{minipage}{0.5\columnwidth}
    \centering
        \includegraphics[width=0.8\columnwidth]{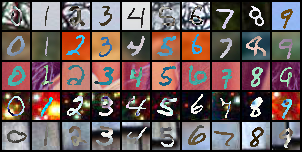}
        \includegraphics[width=0.9\columnwidth]{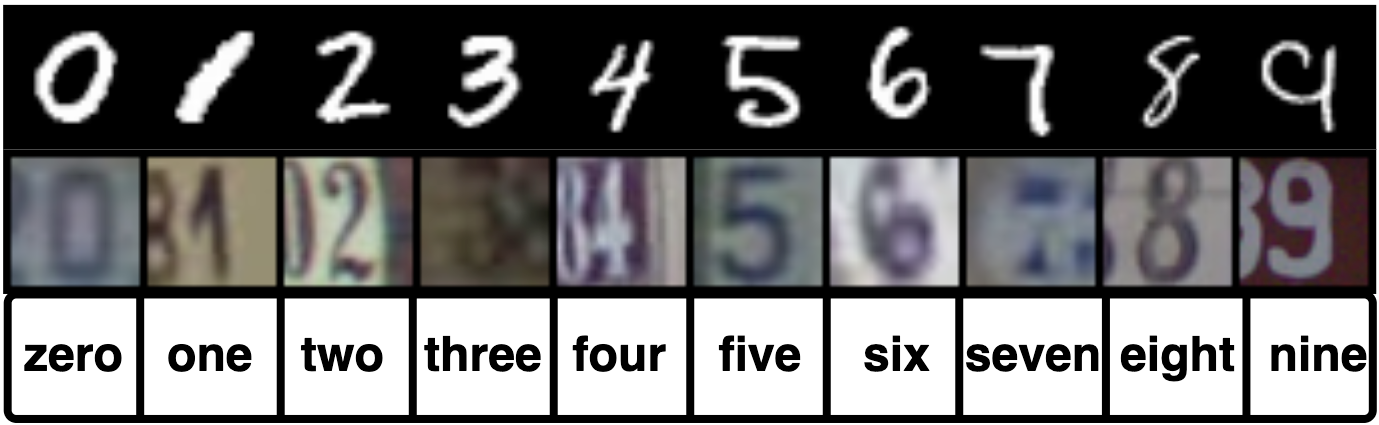}
        % \includegraphics[width=0.19\columnwidth]{cub_data.png}
    % \captionof{figure}{Unconditional joint (top) and conditional cross-modal (bottom) generation on CUB. Dashed boxes denote cross-modal generated results.}
    % \label{fig.cross_gen_cub}
    \end{minipage}
    \begin{minipage}{0.24\columnwidth}
    \centering
        \includegraphics[width=0.9\columnwidth]{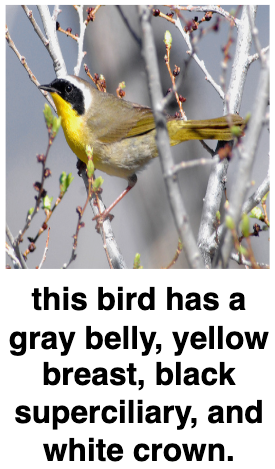}
    \end{minipage}
    \begin{minipage}{0.24\columnwidth}
    \centering
        \includegraphics[width=0.93\columnwidth]{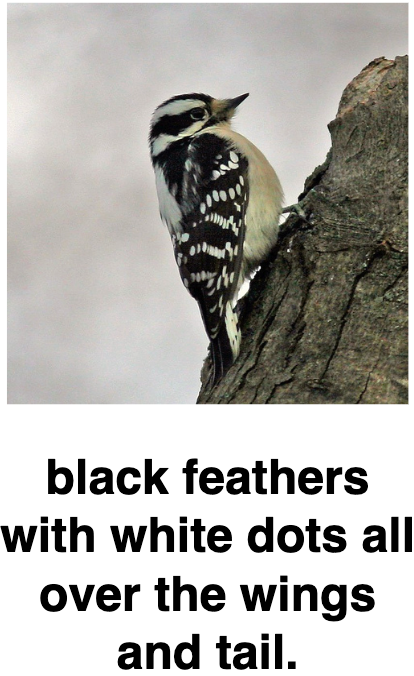}
    \end{minipage}
\captionof{figure}{Example samples from PolyMNIST (top left), MST (bottom left), and CUB (right).}
\label{fig.data_sample}
\end{center}
\begin{figure*}[!t]
    \centering
    \begin{subfigure}[t]{0.49\textwidth}
        \centering
        \includegraphics[width=0.95\textwidth]{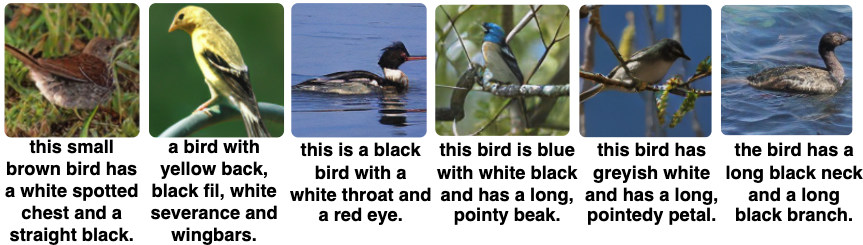}
        % \caption{ShaLa}
    \end{subfigure}%
    \begin{subfigure}[t]{0.49\textwidth}
        \centering
        \includegraphics[width=0.99\textwidth]{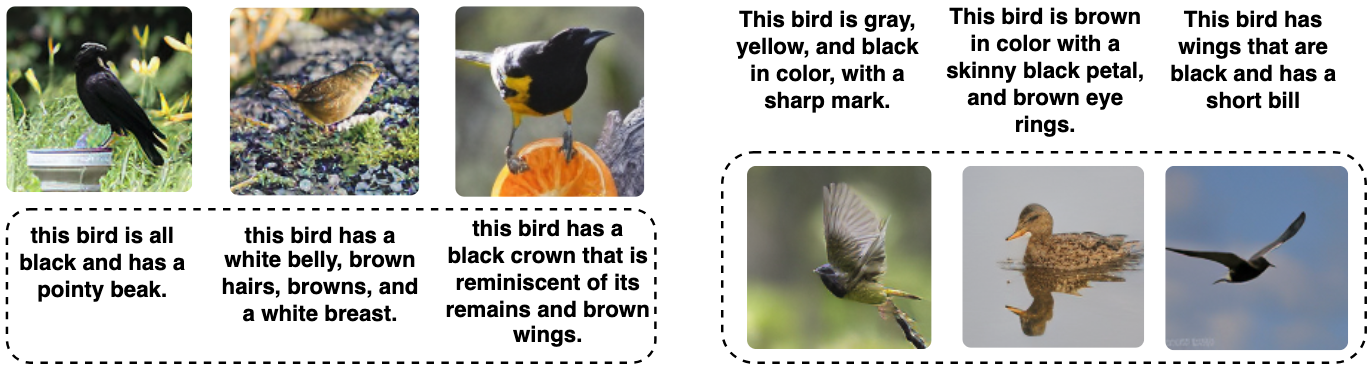}
        % \caption{ShaLa}
    \end{subfigure}%
    % \vspace{1pt}
    \caption{Unconditional (left) and conditional (right) generation on CUB. Dashed boxes denote cross-modal generated results.}
    \label{fig.cross_gen_cub}
\end{figure*}
\subsection{Shared Latent for Synthesis Coherence}\label{sec-multi-syn-coh}
A central objective of shared latent variable generative models is to render semantically coherent outputs across multiple modalities. This includes both unconditional joint generation, i.e., sampling all modalities from the latent prior, and conditional cross-modal inference, i.e., generating target modalities from given modalities. 

\noindent\textbf{Unconditional Joint Coherence.}
We assess ShaLa's capacity for coherent multimodal synthesis on the two benchmarks, PolyMNIST and MST, widely adopted in our direct baselines which use quantitative metrics computed using pre-trained modality-specific classifiers \cite{sutter2021generalizedMoPoE}.
%\footnote{provided in \url{https://github.com/thomassutter/MoPoE}}.
%\noindent\textbf{Unconditional Joint Coherence.} 
%We begin by evaluating ShaLa's ability to generate all modalities simultaneously from samples drawn from the learned latent prior. To measure the semantic coherence of the generated multimodal samples, we employ pre-trained classifiers \cite{sutter2021generalizedMoPoE}. 
These classifiers assess whether the different modalities generated from the same latent code correspond to the same semantic class. Higher coherence indicates a better alignment of modalities in the latent space and thus a more accurate modeling of the joint distribution. 

Unconditional generation forms the bedrock of generative models, and reveals the quality of the shared latent. We evaluate ShaLa's ability to generate all modalities simultaneously from samples drawn from the learned latent prior $\rvz$. As shown in Tab.\ref{tab.coherence_poly_mst}, ShaLa demonstrates superior coherence in this setting, consistently outperforming our baselines.
\begin{table}[!t]
    \centering
    \caption{Coherence ($\uparrow$) for unconditional and conditional cross-modal generation. Note that several baselines do not report results on the MST benchmark. We nonetheless include these models to present a broad landscape of multimodal generative approaches.}
    \resizebox{0.95\columnwidth}{!}{
    \begin{tabular}{c || c  c  c  c}
    \toprule
    \toprule
    & \multicolumn{2}{c}{PolyMNIST} & \multicolumn{2}{c}{MST}\\
    Method & Unconditional & Conditional & Unconditional & Conditional\\
    \midrule
    MVAE  & 0.008 & 0.298 & 0.12 & 0.27\\
    % \hline
    MVTCAE& 0.003 & 0.591 & - & -\\
    % \hline
    mmJSD & 0.060 & 0.778 & - & 0.72\\
    % \hline
    MoPoE & 0.141 & 0.720 & 0.31 & 0.69\\
    MMVAE & 0.232 & 0.844 & 0.28 & 0.68\\
    MMVAE+& 0.344 & 0.869 &  - & -\\
    MVEBM$^*$ & 0.735 & 0.857 & 0.42 & 0.43\\
    CMVAE & 0.781 & \textbf{0.897} & - & -\\
    \textbf{Ours} & \textbf{0.815} & \textbf{0.897} & \textbf{0.44} & \textbf{0.75}\\
    \bottomrule
    \end{tabular}
    }
\label{tab.coherence_poly_mst}
\end{table}

\noindent\textbf{Conditional Cross-Modal Coherence.} We next assess ShaLa in conditional generation, where one modality is used to infer a latent representation that is then used to generate unseen modalities. Specifically, given an input $\rvx_i$, we encode it into a deterministic representation $\rvh_i$ and use it to condition the diffusion-based prior, which samples the latent variable $\rvz$. This latent is then decoded into the target modality $\rvx_j$. The semantic coherence of $\rvx_j$ is again measured using a pre-trained classifier to determine whether it aligns with the semantic label of the observed input $\rvx_i$.

ShaLa consistently achieves high coherence across all modality pairs, reflecting strong alignment and semantic consistency between observed and generated modalities. This effectiveness is largely due to ShaLa's conditioning mechanism, which enables flexible cross-modal generation by leveraging deterministic modality-specific embeddings within a learned diffusion prior. 

\subsection{Shared Latent for Synthesis Quality}\label{sec-exp_multi_syn}
Besides multimodal coherence, a major challenge in variational frameworks is the prior-hole problem, where latent samples drawn from the prior often fall into low-density regions of the aggregated posterior. This misalignment leads to degraded sample quality and unrealistic generations.

ShaLa addresses this issue through the integration of a second-stage diffusion-based latent prior, bridging the distributional gap between the learned posterior and the assumed prior. To assess the learned latent prior, we evaluate the quality of the data generated from this shared latent using  PolyMNIST and CUB, which is considered a more complex and realistic multimodal benchmark. 

We report Fréchet Inception Distance (FID) as our primary quantitative metric. As shown in Tab.\ref{tab.fid_poly_cub}, ShaLa achieves lower FID scores compared to competitive baselines of multimodal shared representation learning, indicating superior sample quality. Qualitative results on CUB in Fig.\ref{fig.cross_gen_cub} illustrate that ShaLa's shared latent can generate visually realistic and semantically meaningful multimodal samples in both unconditional and conditional settings.

These results demonstrate that ShaLa achieves high perceptual fidelity, validating the effectiveness of its diffusion-based latent prior in overcoming key limitations of traditional variational approaches. We report the best results as originally published in our baselines to ensure consistency. 
\begin{table}[!t]
    \centering
    \caption{Generation quality by FID $(\downarrow)$ for unconditional joint and cross-modal generation on PolyMNIST and CUB.}
    \resizebox{0.85\columnwidth}{!}{
    \begin{tabular}{c || c  c  c}
    \toprule
    \toprule
    &  \multicolumn{2}{c}{PolyMNIST} & CUB \\
    Method & Unconditional & Conditional & Conditional\\
    \midrule
    MVAE  & 50.65 & 82.59 & 172.21\\
    % \hline
    MVTCAE& 85.43 & 58.98 & 208.43\\
    % \hline
    mmJSD & 179.76 & 178.27 & 262.80\\
    % \hline
    MoPoE & 98.56 & 160.29 & 265.55\\
    MMVAE & 164.29 & 150.83 & 232.20\\
    MMVAE+& 86.64 & 80.75 & 164.94\\
    MVEBM & - & - & 136.16\\
    CMVAE & 78.52 & 74.53 & 28.00\\
    \textbf{Ours} & \textbf{47.30} & \textbf{40.18} & \textbf{25.58} \\
    \bottomrule
    \end{tabular}
    }
\label{tab.fid_poly_cub}
\end{table}

\subsection{Scalability for Multi-View Challenge}\label{sec-exp_multi_view}
% \begin{center}
%     \begin{minipage}{1.\columnwidth}
%     \centering
%     \includegraphics[width=0.85\columnwidth]{shapenet_data.png}
%     \captionof{figure}{Example 16-view samples from ShapeNet Car.}
%     \label{fig.shapnet_data}
%     \end{minipage}
% \end{center}
To further assess the scalability of ShaLa, we evaluate its performance on multi-view datasets, where different viewpoints of the same object are treated as distinct modalities. Specifically, we conduct experiments on ShapeNet \cite{shapenet2015}, a widely used benchmark for view-consistent synthesis. In this setting, each view captures complementary visual information of the same 3D object, analogously to how multimodal data share the same semantic content.

\noindent\textbf{Experiment Setting.} We frame this task as a challenging instance of multimodal generative modeling, due to the significantly increased number of modalities (i.e., views). Following standard practice \cite{liu2023syncdreamer, anciukevivcius2023renderdiffusion}, we render 16 canonical views per object, each resized to a resolution of 128$\times$128. We assess cross-view synthesis quality, i.e., generating unseen views, using two standard image similarity metrics: peak-signal-to-noise-ratio (PSNR) and structural-similarity-index-measure (SSIM). 

\begin{center}
    \begin{minipage}{0.99\columnwidth}
    \centering
    \includegraphics[width=1\columnwidth]{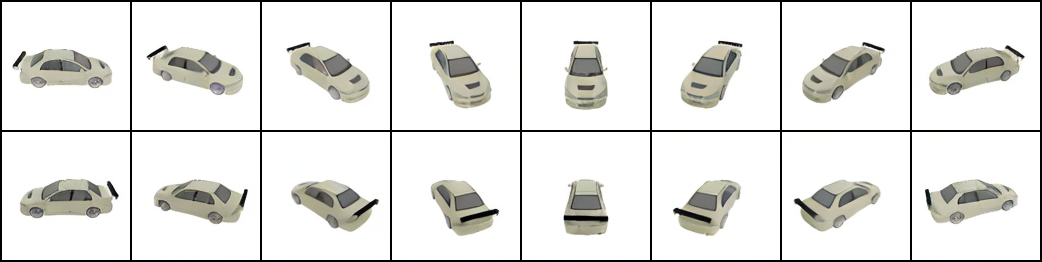}
    \includegraphics[width=1\columnwidth]{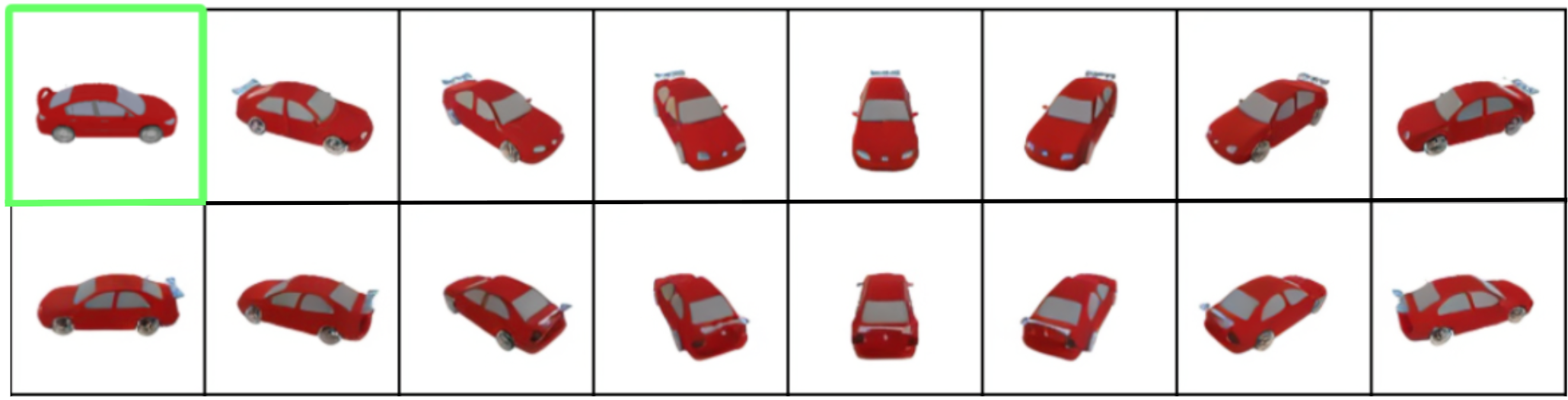}
    \captionof{figure}{ShaLa unconditional and cross-modal multi-view generation. The input views are denoted by green boxes.}
    \label{fig.ours_car_results}
    \end{minipage}
\end{center}
\begin{figure*}[!t]
    \centering
    \includegraphics[width=0.9\textwidth]{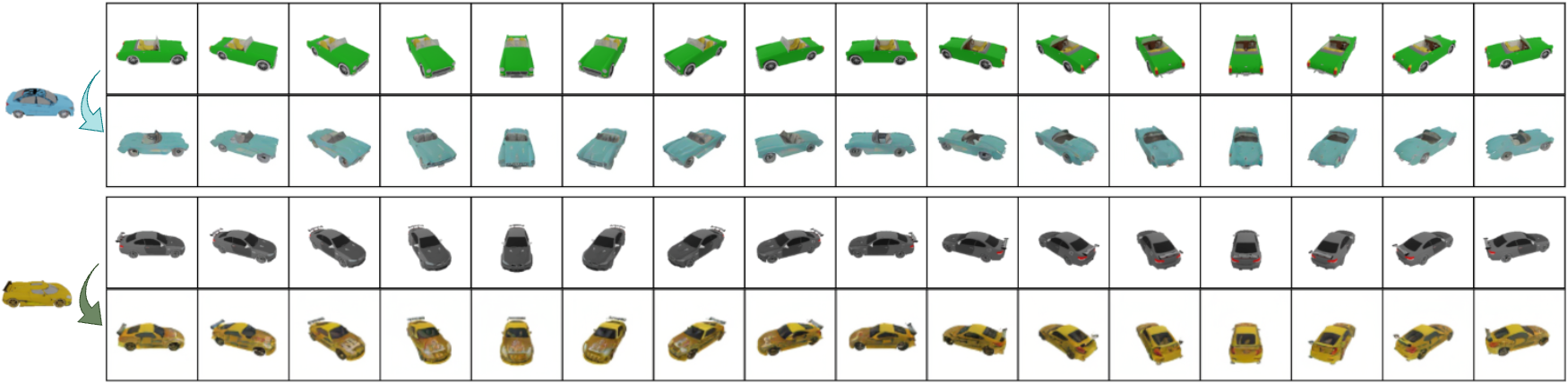}
    \caption{Visualization of latent transfer. The left column is $\rvX^{\text{ref}}$, the first row is $\rvX^{\text{source}}$, and the second row is the result.}
    \label{fig.style}
\end{figure*}
\noindent\textbf{Benchmark performance.} We adapt MMVAE+ \cite{palumbo2023mmvae+} and CMVAE \cite{palumbo2024deepcmvae}, both of which serve as prominent shared latent variable backbones in multimodal VAEs. As a reference, we also report results from \textit{task-specific view synthesis baselines}, including SyncD \cite{liu2023syncdreamer}, Px-NeRF \cite{yu2021pixelnerf}, EG3D \cite{chan2022efficienteg3d}, and RenderD \cite{anciukevivcius2023renderdiffusion}. These methods incorporate \textit{explicit 3D priors or inductive biases} to maintain geometric consistency and are tailored for the view synthesis task. In contrast, ShaLa is trained \textit{without any 3D supervision or assumptions}, relying solely on image views and latent alignment. As shown in Tab.\ref{tab.psnr_car}, ShaLa significantly outperforms MMVAE+ and CMVAE. Compared to task-specific methods, it achieves competitive performance, highlighting the flexibility and effectiveness of ShaLa as a unified generative modeling framework for both traditional multimodal and multi-view data.
\begin{center}
    \begin{minipage}{0.99\columnwidth}
    \centering
    \captionof{table}{PSNR ($\uparrow$) and SSIM ($\uparrow$) for novel view synthesis. Ours$^*$ use image size 256$\times$256 as SyncD for fairness.}
    \resizebox{0.99\columnwidth}{!}{
    \begin{tabular}{c || c c c | c  c  c  || c  c}
    \toprule
    \toprule
     & Ours & MMVAE+ & CMVAE & Px-NeRF & EG3D  & RenderD & Ours$^*$ & SyncD\\
    \midrule
    PSNR & 24.7 & 19.3 & 20.5 & 23.2 & 21.8 & 25.4 & 22.2 & 21.9\\
    SSIM & 0.89 & 0.59 & 0.64 & 0.90 & 0.71 & 0.81 & 0.88 & 0.88\\
    \bottomrule
    \end{tabular}
    }
\label{tab.psnr_car}
\end{minipage}
\end{center}

\subsection{Latent-Based Multi-View Style Transfer}
By modeling a meaningful and flexible shared latent space, ShaLa can support a range of downstream tasks that go beyond traditional joint and conditional generation. In this section, we explore multi-view style transfer, a novel application setting that transfers a reference style to multiviews where the generated views not only carry the style (e.g., texture of car) but also maintain the structure presented in the source views (e.g, shape of car). This can be challenging for prior multimodal VAEs lack a conditioning mechanism, or task-specific models such as SyncD, which are typically constrained to only novel-view synthesis.

Given a source instance $\rvX^{\text{source}}$ and a reference instance $\rvX^{\text{ref}}$, each comprising multiple views, the goal is to transfer low-level visual features (e.g., color, texture) from the reference to the source while preserving the source's structural semantics (e.g., shape). We achieve this by first sampling a latent code $\rvz^{\text{source}} \sim q_\phi(\rvz|\rvX^{\text{source}})$ and applying $K$ steps of forward diffusion to obtain a partially perturbed latent $\rvz^{\text{source}}_K$. This serves to remove view-specific low-level information while retaining higher-level semantic structure. Then, we condition the reverse diffusion process on a deterministic feature $\rvh_j^{\text{ref}}$ randomly selected from $\rvX^{\text{ref}}$ to resample the latent code for the decoder to synthesize outputs
%that preserve the global structure of the source 
while adopting style features from the reference.

Qualitative results shown in Fig.\ref{fig.style} demonstrate ShaLa's ability to produce faithful structural preservation with successful style transfer across views. These results highlight the expressive and compositional nature of the shared latent representation learned by ShaLa. 

\subsection{Ablation Study}\label{sec-ablation}
\noindent\textbf{Architectural Inference.} In Eqn.\ref{our-inf}, our architectural inference model encodes each modality $\rvx_i$ into a deterministic embedding $\rvh_i$, which is then fused to produce a shared latent posterior. This formulation treats $\rvh_{1:m}$ as an information bottleneck, providing a compact and semantically meaningful conditioning signal for the second-stage diffusion prior. To validate this, we compare performance when conditioning the diffusion model directly on $\rvx_{1:m}$ versus on $\rvh_{1:m}$. As shown in Tab.\ref{table.conditional-x}, using $\rvh_{1:m}$ improves both unconditional and conditional synthesis, suggesting the benefit of our architectural inference model.

\begin{center}
    \begin{minipage}{0.99\columnwidth}
    \centering
    \captionof{table}{Coherence of using different conditional signals.}
    \resizebox{0.8\columnwidth}{!}{
    \begin{tabular}{c || c  c  c  c}
    \toprule
    \toprule
    & \multicolumn{2}{c}{PolyMNIST} & \multicolumn{2}{c}{MST}\\
    Condition & Unconditional & Conditional & Unconditional & Conditional\\
    \midrule
    $\rvx_{1:m}$ & 0.584 & 0.612 & 0.23 & 0.35\\
    $\rvh_{1:m}$ & \textbf{0.815} & \textbf{0.897} & \textbf{0.44} & \textbf{0.75}\\
    \bottomrule
    \end{tabular}
    }
\label{table.conditional-x}
\end{minipage}
\end{center}

\noindent\textbf{Ablation of Diffusion Prior.} We assess the effect of the diffusion prior by varying the capacity of the diffusion network. As shown in Tab.\ref{table.varying}, reducing the model size leads to noticeable drops in generation quality (as measured by FID), while increasing the model size yields consistent improvements. This suggests that ShaLa’s shared latent space provides a stable foundation across different diffusion configurations, and that our second-stage diffusion prior effectively leverages this representation for high-quality synthesis.

\begin{center}
    \begin{minipage}[!h]{0.99\linewidth}
    \centering
    \captionof{table}{Varying diffusion configuration.}
    \label{table.varying}
    \resizebox{0.7\columnwidth}{!}{
    \begin{tabular}{c|c|c|c|c}
        \toprule
        \toprule
        FID / \#P (parameter) & 2$\times$ \#P & 1$\times$ \#P & 1/2$\times$ \#P & 1/4$\times$ \#P \\
        \midrule
        Unconditional & \textbf{45.24}  & 47.30 & 50.24 & 56.44 \\
        Conditional   & \textbf{37.15}  & 40.18 & 45.86 & 49.75 \\
        \bottomrule
    \end{tabular}
    }
\end{minipage}
\end{center}
\noindent\textbf{Variation of Fuse Function.} In our default implementation, the fusion function $\mathbf{\hbar} = \odot(\rvh_1, \dots, \rvh_M)$ is realized by concatenating the modality-specific embeddings and applying four layers of linear transformations, i.e., $\mathbf{\hbar} = F([\rvh_1, \dots, \rvh_M])$. This design empirically achieves strong performance. To assess the impact of this choice, we explore two alternative fusion strategies:  (i) Summation-based Fusion $\mathbf{\hbar} = F(\sum_{i=1}^M\rvh_i)$ and (ii) Gated Fusion $\mathbf{\hbar} = F([g_1(\rvh_1), \dots, g_M(\rvh_M)])$, where each modality-specific embedding is first transformed via a learnable gating function $g_i$, followed by concatenation and shared fusion. On PolyMNIST, the default concatenation-based fusion achieves better conditional coherence, while the summation and gated fusion variants result in lower coherence scores (0.75 and 0.64, respectively).

\noindent\textbf{Number of Modalities.} Scalability to a larger number of modalities is critical for shared latent variable models. To assess this, we examine ShaLa with advanced multimodal VAEs, such as CMVAE and MMVAE$+$, with increasing number of modalities. As shown in Tab. \ref{table.psnr-cmvae}, ShaLa maintains highest PSNR and SSIM, whereas the baselines degrade with more views. This demonstrates the robustness and scalability of ShaLa in modeling complex multimodal distributions. 
\begin{center}
    \begin{minipage}{0.99\columnwidth}
    \centering
    \captionof{table}{Comparison of using different conditional signals.}
    \resizebox{0.8\columnwidth}{!}{
    \begin{tabular}{c || c  c  c| c  c  c}
    \toprule
    \toprule
     & \multicolumn{3}{c}{3 views} & \multicolumn{3}{|c}{8 views} \\
     & Ours & MMVAE+ & CMVAE & Ours & MMVAE+ & CMVAE \\
    \midrule
    PSNR $\uparrow$ & \textbf{26.4} & 22.5 & 23.3 & \textbf{25.6} & 20.2 & 21.8 \\
    SSIM $\uparrow$ & \textbf{0.94} & 0.67 & 0.72 & \textbf{0.92} & 0.61 & 0.68 \\
    \bottomrule
    \end{tabular}
    }
\label{table.psnr-cmvae}
\end{minipage}
\end{center}
\section{Conclusion}
In this work, we present a novel framework for learning shared latent space, coined as ShaLa. We leverage an architectural inference model for learning the shared latent variables, where the conditional deterministic variables can facilitate the second-stage shared latent diffusion model. Extensive experiments demonstrate favorable results of ShaLa and the scalability in modelling the difficult multi-view data.

\appendix
\section{Additional Ablation Study and Experiment}
\begin{figure*}[!t]
    \centering
    \begin{subfigure}[t]{0.95\textwidth}
        \centering
        \includegraphics[width=0.99\textwidth]{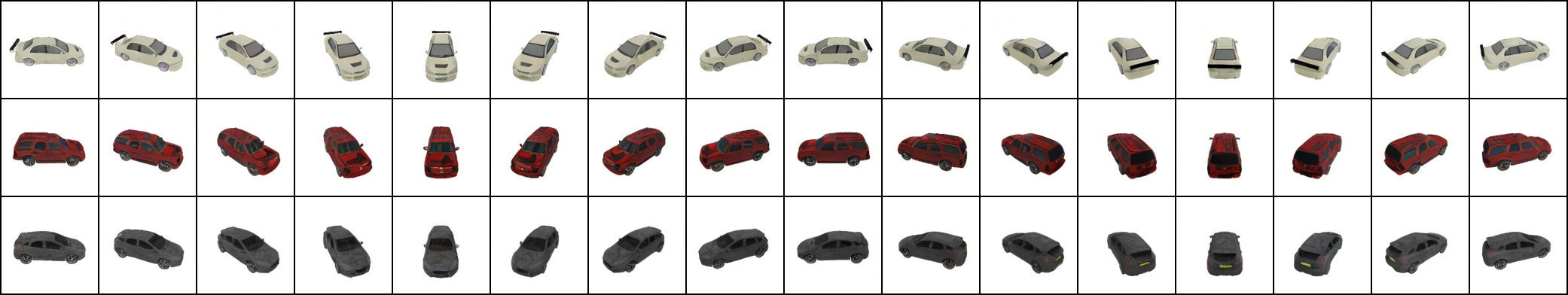}
        \caption{Unconditional joint multi-view generation.}
        \label{fig.joint_car}
    \end{subfigure}%
    
    \begin{subfigure}[t]{0.95\textwidth}
        \centering
    \includegraphics[width=0.99\textwidth]{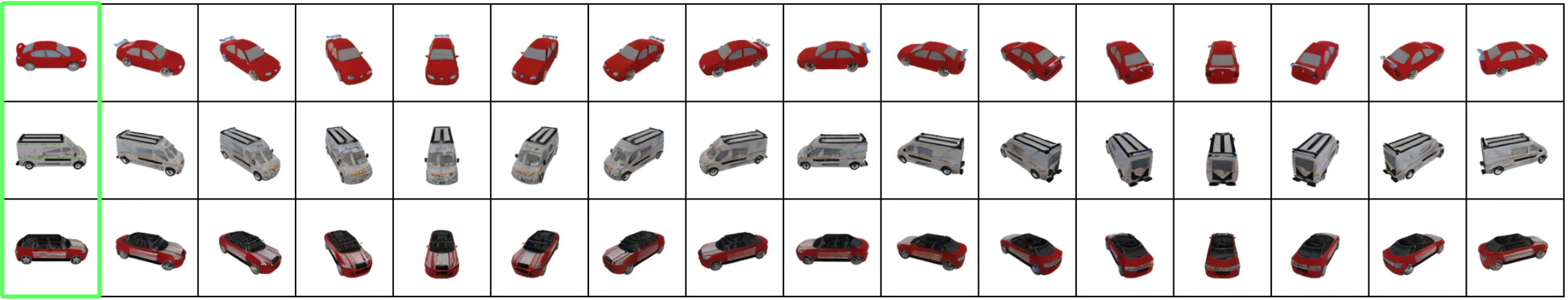}
    \caption{Cross-modal multi-view generation (novel-view generation). The input views are denoted by green boxes.}
    \label{fig.cross_car}
    \end{subfigure}
    \caption{Qualitative results of ShaLa on ShapeNet Car.}
    \label{fig.joint_cross_car}
\end{figure*}

\begin{center}
    \begin{minipage}{1.\columnwidth}
    \centering
    \includegraphics[width=0.99\columnwidth]{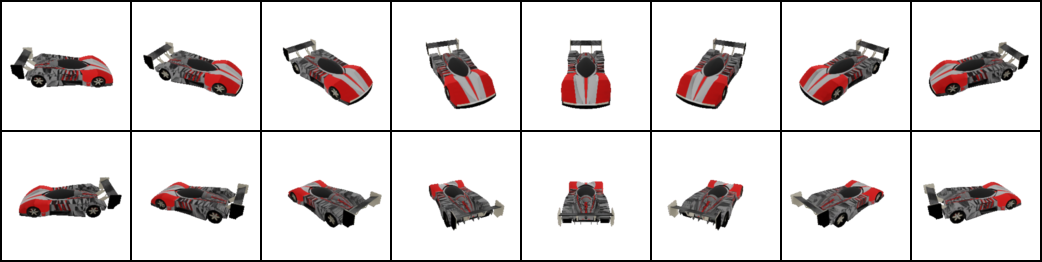}
    \captionof{figure}{Data example 16-view samples for ShapeNet Car.}
    \label{fig.shapnet_data}
    \end{minipage}
\end{center}
\begin{center}
    \begin{minipage}{0.99\columnwidth}
    \centering
    \includegraphics[width=0.99\columnwidth]{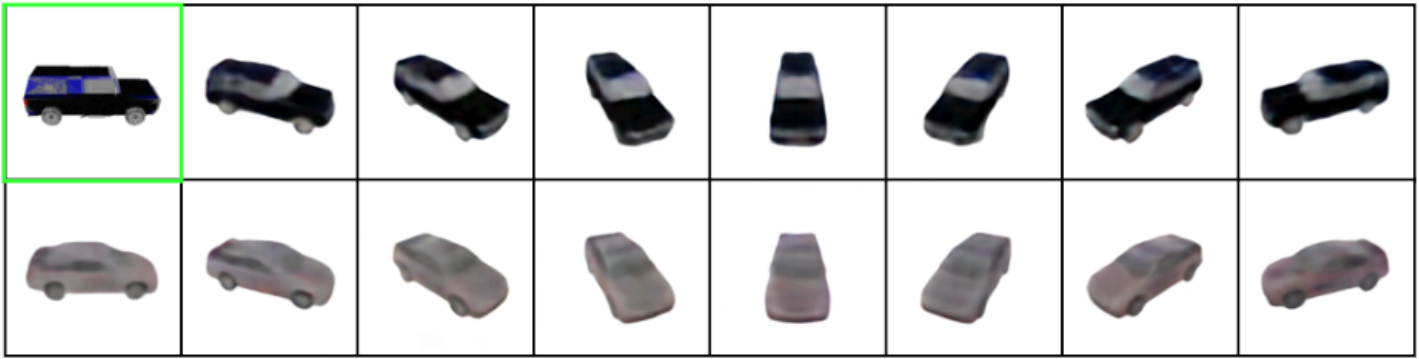}
    \captionof{figure}{Condition (input denoted by green box) and unconditional synthesis for CMVAE for 8-view. See comparison to our 16-view results in Fig. \ref{fig.joint_cross_car}.}
    \label{fig.shapnet_cmvae}
    \end{minipage}
\end{center}
\subsection{Latent-based Multi-View Correction}\label{sec-app-correct}
We further investigate ShaLa's ability to recover from incomplete or corrupted multimodal inputs through a process we refer to as latent correction. This task differs fundamentally from conditional generation: rather than inferring a latent code from a reliable subset of modalities, we begin with a potentially abnormal latent code derived from corrupted inputs and attempt to correct it via resampling.

Suppose we are given a subset $\hat{\rvX} = {\rvx_i, \dots, \rvx_j}$ containing only partial or corrupted observations. Direct encoding from this subset yields an ill-defined latent $\hat{\rvz} \sim q_\phi(\rvz|\hat{\rvX})$, resulting in incoherent or collapsed reconstruction. To correct this, we apply $K$ steps of the forward diffusion process to add noise and partially destroy malicious features. We then perform reverse diffusion to resample a corrected latent code.
\begin{figure*}[!t]
    \centering
    \includegraphics[width=0.99\textwidth]{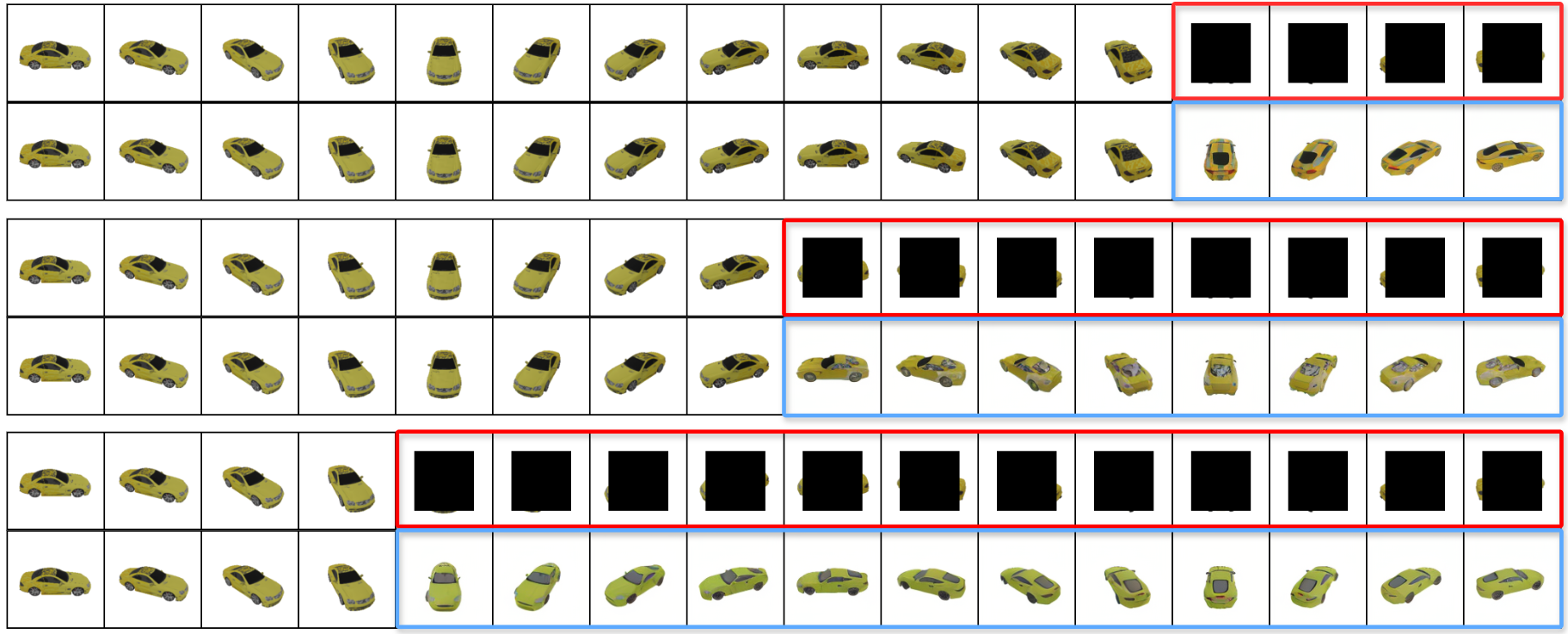}
    \caption{Latent correction. The broken and corrected modalities are denoted by red boxes and blue boxes, respectively.}
    \label{fig.inpainting}
\end{figure*}
Fig. \ref{fig.inpainting} shows that ShaLa is able to restore coherent and semantically valid generations, even as the number of corrupted modalities increases. These results demonstrate that ShaLa's learned diffusion prior not only enables high-quality generation but also provides a mechanism for robust inference in challenging, degraded input settings.

\section{Derivation and Algorithm}\label{sec-app-tec-alg}
Recall that the shared latent variable model is formalized as a joint distribution, i.e., $p_{\theta}(\rvX, \rvz)=p_{\theta}(\rvX|\rvz)p_0(\rvz)$. Given $N$ multimodal examples $\{\rvX_i\}_{i=1}^N$, we maximize the log-likelihood, i.e., $\max_\theta \mathcal{L}(\theta)=\frac{1}{n}\sum_{i=1}^N\log p_\theta(\rvX_i)$. In the asymptotic regime $N \rightarrow \infty$, this is equivalent to minimizing the KL divergence between the true data distribution and the model distribution, i.e., $\mathrm{KL}(\p(\rvX) || p_\theta(\rvX))$. The gradient $\frac{\partial}{\partial \theta}\mathcal{L}(\theta)$ is computed as
\begin{equation}\label{mle-grad}
\begin{aligned} 
    &\frac{\partial}{\partial \theta}\log p_{\theta}(\rvX) = \E_{p_\theta(\rvz|\rvX)}[\frac{\partial}{\partial \theta}\log p_\theta(\rvX)]\\
    &=\E_{p_\theta(\rvz|\rvX)}[\frac{\partial}{\partial \theta}\log p_\theta(\rvX)] + \E_{p_\theta(\rvz|\rvX)}[\frac{\partial}{\partial \theta}\log p_\theta(\rvz|\rvX)]\\
    &=\E_{p_\theta(\rvz|\rvX)}[\frac{\partial}{\partial \theta}\log p_\theta(\rvX, \rvz)]
\end{aligned}
\end{equation}
where the first equation is a simply identity, and the second equation follows from $\E_{p_\theta(\rvz|\rvX)}[\frac{\partial}{\partial \theta}\log p_\theta(\rvz|\rvX)]=\int p_\theta(\rvz|\rvX)[\frac{\partial}{\partial \theta}\log p_\theta(\rvz|\rvX)] d\rvz=\frac{\partial}{\partial \theta}\int p_\theta(\rvz|\rvX)d\rvz=0$.

By introducing a parameterized inference model $q_\phi(\rvz|\rvX)$ (we omit specific factorizations for simplicity), learning both the inference and generative model can be achieved by minimizing a joint KL divergence as $\mathrm{KL}(\p(\rvX)q_\phi(\rvz|\rvX) || p_\theta(\rvX, \rvz))$, which can be viewed as a surrogate of the MLE objective with the KL perturbation term,
\begin{equation}\label{var-grad-elbo}
\begin{aligned} 
    &\mathrm{KL}(\p(\rvX)q_\phi(\rvz|\rvX) || p_\theta(\rvX, \rvz))\\
    = &\mathrm{KL}(\p(\rvX) || p_\theta(\rvX)) + \mathrm{KL}(q_\phi(\rvz|\rvX) || p_\theta(\rvz|\rvX))
\end{aligned}
\end{equation}
The second KL term measures the KL-divergence between the inference model and the true generator posterior, and by minimizing it, the inference model is learned to approximate the true generator posterior, forming a tractable surrogate learning objective. Specifically, the gradient $\frac{\partial}{\partial \theta,\phi}\mathcal{L}(\theta,\phi)$ is computed as 
\begin{equation}\label{var-grad}
\begin{aligned} 
    \frac{\partial}{\partial \theta,\phi}\mathcal{L}(\theta,\phi)=\frac{\partial}{\partial \theta,\phi}\E_{q_\phi(\rvz|\rvX)}[\log \frac{p_\theta(\rvX, \rvz)}{q_\phi(\rvz|\rvX)}]
\end{aligned}
\end{equation}
In multimodal VAEs, the specific structure of $q_\phi(\rvz|\rvX)$ varies across models, often involving factorized or aggregated unimodal posteriors (e.g., PoE, MoE). These factorizations typically require parallel computation across modality-specific encoders and sub-sampled training strategies. In contrast, our approach parameterizes the joint posterior $q_\phi(\rvz|\rvX)$ directly as a single Gaussian distribution inferred from fused deterministic representations. This enables efficient and unified training without the overhead of computing or combining modality-specific posterior distributions.

For our diffusion prior, we follow DDPM \cite{ho2020denoising} for the training algorithm but instead use our designed reverse kernel (Eqn. 7 in the paper) for modelling. We refer to the detailed derivation of the diffusion learning objective in the original paper.

\noindent\textbf{Implementations.} For all experiments, we implement our architectural inference model using $M$ encoder-decoder pairs, where each encoder consists of multiple convolutional and down-sampling layers, and each decoder mirrors this structure with up-sampling operations. To construct the fused representation, we concatenate the deterministic features from each modality and apply a four-layer feedforward network to produce the final fused vector $\hbar$.

Our latent diffusion prior follows the standard denoising diffusion probabilistic modeling (DDPM) framework. We adopt a diffusion transformer as the core network, which empirically provides strong modeling capacity and robust inference behavior. The model is trained using a time-averaged objective with randomly sampled diffusion steps, following common diffusion practices. 

All models are trained on A100 Nvidia GPU using common programming tool (e.g., Pytorch), standard optimization techniques, and evaluated with consistent protocols across datasets. Full architectural details and training hyperparameters will be released upon acceptance to facilitate reproducibility.

\section{Additional Qualitative Result}\label{sec-app-add}
\begin{center}
    \begin{minipage}{0.99\columnwidth}
    \centering
    \includegraphics[width=0.49\columnwidth]{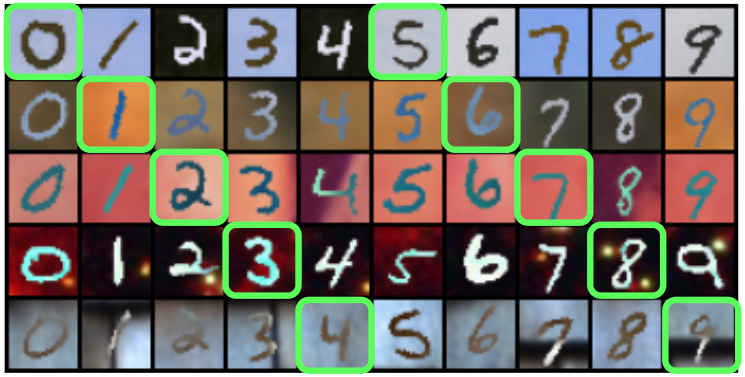}
    \includegraphics[width=0.49\columnwidth]{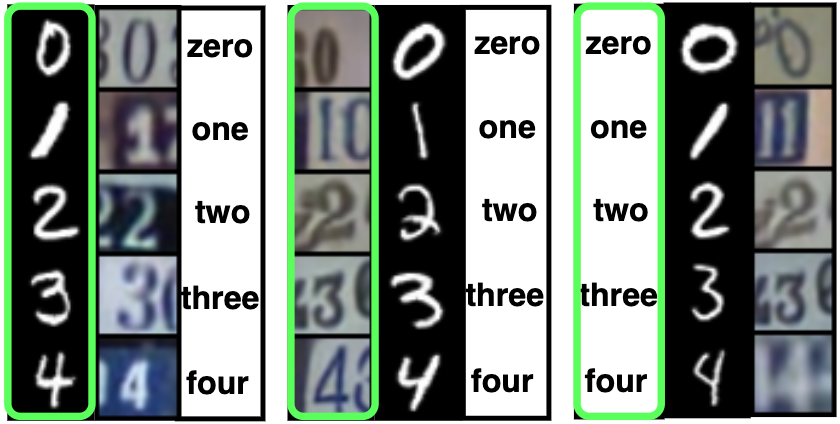}
    \captionof{figure}{Cross-modal generation on PolyMNIST (\textbf{left}) and MST (\textbf{right}). Input modality is denoted by green boxes.}
    \label{fig.ours_poly_results-app}
    \end{minipage}
\end{center}
\begin{figure*}[!t]
    \centering
    \begin{subfigure}[t]{0.199\textwidth}
        \centering
        \includegraphics[width=0.99\textwidth, height=0.75in]{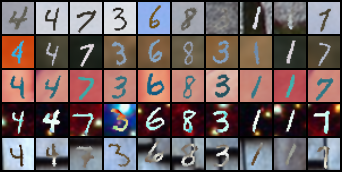}
        \caption{ShaLa}
    \end{subfigure}%
    % \\
    \begin{subfigure}[t]{0.199\textwidth}
        \centering
        \includegraphics[width=0.99\textwidth, height=0.75in]{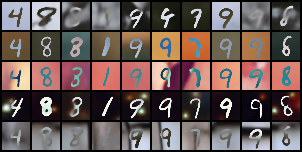}
        \caption{CMVAE}
    \end{subfigure}%
    % \\
    \begin{subfigure}[t]{0.199\textwidth}
        \centering
        \includegraphics[width=0.99\textwidth, height=0.75in]{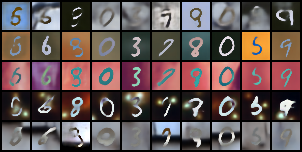}
        \caption{MMVAE$+$}
    \end{subfigure}%
    % \\
    \begin{subfigure}[t]{0.199\textwidth}
        \centering
        \includegraphics[width=0.99\textwidth, height=0.75in]{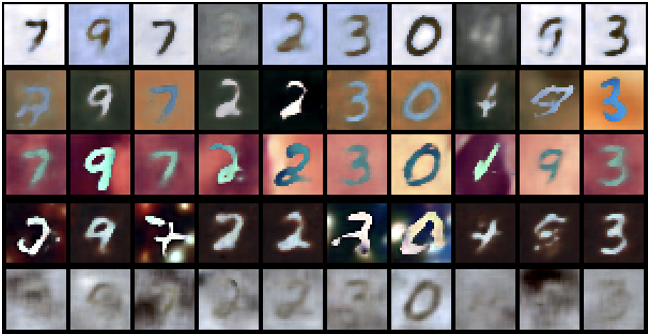}
        \caption{MVEBM}
    \end{subfigure}%
    % \\
    \begin{subfigure}[t]{0.199\textwidth}
        \centering
        \includegraphics[width=0.99\textwidth, height=0.75in]{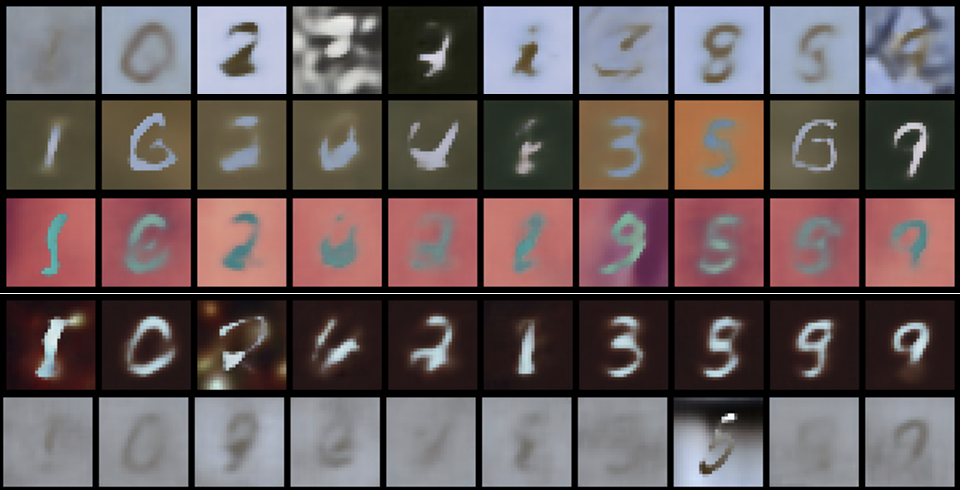}
        \caption{MMVAE}
    \end{subfigure}%
    \caption{Qualitative comparison of unconditional joint generation on PolyMNIST.}
    \label{fig.joint_gen_poly_cub}
\end{figure*}

Fig. \ref{fig.ours_poly_results-app} presents ShaLa’s conditional cross-modal generation results for both PolyMNIST and MNIST-SVHN-Text datasets. The input modality used for conditioning is highlighted in green. As illustrated in Fig. \ref{fig.joint_gen_poly_cub}, ShaLa also outperforms all baseline methods in unconditional generation tasks, consistently producing higher-quality and more semantically coherent outputs.

\section{Limitation and Future Work}
\noindent\textbf{Fine-grained Control.} While ShaLa demonstrates strong performance in generating semantically coherent multimodal content, its ability to control fine-grained details remains limited. The shared latent variable, combined with conditioning via deterministic features, may not offer sufficient precision for tasks that demand localized or high-frequency control—such as detailed texture editing or object-level manipulation. This limitation is inherent in the coarse granularity of the shared latent space and is a common challenge across latent diffusion-based generative frameworks.

\noindent\textbf{Inference Overhead.} Compared to Multimodal VAEs, ShaLa samples latents from diffusion prior, which requires iterating through the backward trajectory, introducing additional inference overhead. Specifically, we measure inference cost using NFE (number of function evaluations), as the practical platform (e.g., GPU type, language efficiency) can vary widely. ShaLa requires 251 NFE (number of denoising steps and decoding), whereas some Multimodal VAEs (e.g., MMVAE+) only need 1 NFE (decoding). However, the overhead of ShaLa is manageable with a growing number of modalities because the diffusion model operates in one shared latent space of all modalities, rather than separate operations on the data or latents of each modality.

\noindent\textbf{Generalization to Unseen Modalities.}
ShaLa is trained on fixed sets of modalities and is not explicitly designed to accommodate unseen or novel modality types during inference. While the architectural inference model generalizes well to modality dropout within the training set, extending ShaLa to support open-modality settings—such as zero-shot modality transfer—remains an open challenge. Future work could explore incorporating modality-agnostic embeddings or meta-learning strategies to improve generalization beyond the training distribution.

\begin{algorithm}[!]
   \caption{ShaLa training algorithm}
   \label{alg}
\begin{algorithmic}
   \STATE {\bfseries Input:} Training iterations for the first and second stage $T_1$, $T_2$, observed training examples $\{\rvX_i\}_{i=1}^N$, batch size $b$, network parameters $\theta, \phi, \beta$, learning rate $\eta_{\theta}, \eta_{\phi}, \eta_{\beta}$, 
   \STATE Let $t = 0$;
   \REPEAT
   \STATE \textbf{Update $\theta$, $\phi$:} Given$\{\rvX_i\}_{i=1}^b$, compute gradient by Eqn. \ref{var-grad} and update with $\eta_{\theta}$ and $\eta_{\phi}$
   
   \STATE Let $t = t + 1$;
   \UNTIL{$t = T_1$}
   \STATE Let $t = 0$;
   \REPEAT
    \STATE \textbf{Sample $\rvh_{1:M}$, $\rvz$:} Given$\{\rvX_i\}_{i=1}^b$, encode $\rvh_{1:M}$ and $\rvz$ from our architectural inference model.
    \STATE \textbf{Random Draw $\rvh_i$:} Sample $\rvh_i \sim \mathrm{Uniform}(\rvh_{1:M})$.
    \STATE \textbf{Update $\beta$:} Use $\rvh_i$, $\rvz$ for our reverse kernel (Eqn. 7 in paper) and compute diffusion learning objective (i.e., Eqn. 14 in \cite{ho2020denoising}).
    \UNTIL{$t = T_2$}
\end{algorithmic}
\end{algorithm}
% \bibliography{aaai2026}

% \end{document}

\newpage
\bibliography{aaai2026}

\end{document}